\RequirePackage[hyphens]{url}

\documentclass[twoside,11pt]{article}

\usepackage{jmlr2e}

\usepackage{amsmath}

\ShortHeadings{To tune or not to tune the number of trees in random forest?}{Probst and Boulesteix}
\firstpageno{1}

\begin{document}

\title{To tune or not to tune the number of trees in random forest?}

\author{\name Philipp\ Probst \email probst@ibe.med.uni-muenchen.de \\
       \addr Institut f{\"u}r medizinische Informationsverarbeitung, Biometrie und Epidemiologie \\
       Marchioninistr. 15, 81377 M{\"u}nchen
       \AND
       \name Anne-Laure\ Boulesteix \email boulesteix@ibe.med.uni-muenchen.de \\
       \addr Institut f{\"u}r medizinische Informationsverarbeitung, Biometrie und Epidemiologie \\
       Marchioninistr. 15, 81377 M{\"u}nchen}

\editor{}

\maketitle

\begin{abstract}
The number of trees T in the random forest (RF) algorithm for supervised learning has to be set by the user. It is controversial whether T should simply be set to the largest computationally manageable value or whether a smaller T may in some cases be better. While the principle underlying bagging is that ”more trees are better”, in practice the classification error rate sometimes reaches a minimum before increasing again for increasing number of trees. The goal of this paper is four-fold: (i) providing theoretical results showing that the expected error rate may be a non-monotonous function of the number of trees and explaining under which circumstances this happens; (ii) providing theoretical results showing that such non-monotonous patterns cannot be observed for other performance measures such as the Brier score and the logarithmic loss (for classification) and the mean squared error (for regression); (iii) illustrating the extent of the problem through an application to a large number (n = 306) of datasets from the public database OpenML; (iv) finally arguing in favor of setting it to a computationally feasible large number, depending on convergence properties of the desired performance measure.
\end{abstract}

\section{Introduction}
The random forest (RF) algorithm for classification and regression, which is based on the aggregation of a large number $T$ of decision trees, was first described in its entirety by \citet{Breiman2001}. $T$ is one of several 
important parameters which have to be carefully chosen by the user. It is controversial, however, whether the number of trees $T$ should simply be set to the largest computationally manageable value or whether a smaller $T$ may in some 
cases be better, in which case $T$ should ideally be tuned carefully. This question is relevant to any user of RF and has been the topic of much informal discussion in the scientific community, but has to our knowledge never been addressed systematically from a theoretical and empirical point of view.

\citet{Breiman2001} provides proofs of convergence for the generalization error in the case of classification random forest for growing number of trees. This means that the error rate for a given test or training dataset 
converges to a certain value. Moreover, \citet{Breiman2001} proves that there exists an upper bound for the generalization error. Similarly he proves the convergence of the mean squared generalization error for regression 
random forests and also provides an upper bound. However, these results do not answer the question of whether the number of trees is a tuning parameter or should be set as high as computationally feasible, although convergence 
properties may at first view be seen as an argument in favor of a high number of trees. Since each tree is trained individually and without knowledge of previously trained trees, the risk of overfitting when adding more trees discussed by 
\citet{Friedman2001} in the case of boosting  is not relevant here. 

These arguments however do not give us confidence that a larger $T$ is always better. As a matter of fact, the number of trees is sometimes considered as a tuning parameter in the literature \citep{Raghu2015}; see also 
\citet{Barman2014} for a study in which different random seeds are tested to obtain better forests. The R package \texttt{RFmarkerDetector}  \citep{Palla2016} even provides a function, 'tuneNTREE', to tune the number of 
trees. Of note, the question of whether a smaller number of trees may be better has often been discussed in online forums (see Supplementary File (at the end of the article) for a non-exhaustive list of links) and seems to remain a controversial issue to date. 

A related but different question is whether a smaller number of trees is \textit{sufficient} (as opposed to \lq\lq better'') in the sense that more trees do not improve accuracy. This question is examined, for example, 
in the very early study by \citet{Latinne2001} or by \citet{Hernandez2013}. Another important contribution to our question is the study by \citet{Oshiro2012}, which compared the performance in terms of the Area Under the ROC Curve (AUC) of random 
forests with different numbers of trees on 29 datasets. Their main conclusion is that the performance of the forest does not always substantially improve as the number of trees grow and after having trained a certain 
number of trees (in their case 128) the AUC performance gain obtained by adding more trees is minimal. The study of \citet{Oshiro2012} provides important empirical support for the existence of a \lq\lq plateau'', 
but does not directly address the question of whether a smaller number of trees may be substantially better and does not investigate this issue from a theoretical perspective, thus making the conclusions dependent on the 29 examined datasets. 

In this context, the goal of our paper is four-fold: (i) providing theoretical results showing that the expected error rate may be a non-monotonous function of the number $T$ of trees and an explanation in which circumstance this happens; (ii) providing theoretical results showing that such a non-monotonous pattern can also be observed for the AUC but not for the probability based measures Brier score and logarithmic loss (for classification) and the mean squared 
error (for regression); (iii) illustrating the extent of the problem through an application to a large number ($n=306$) of datasets from a public database; (iv) finally arguing, based on the theoretical results and empirical 
illustration, against tuning of $T$ in practice or, in other words, in favor of setting it to a computationally feasible large number, depending on convergence properties of the desired performance measure, which can be examined with the help of our new 
R package \texttt{OOBCurve}.

\begin{figure}[!htb]
\begin{center}
  \includegraphics[width=\textwidth]{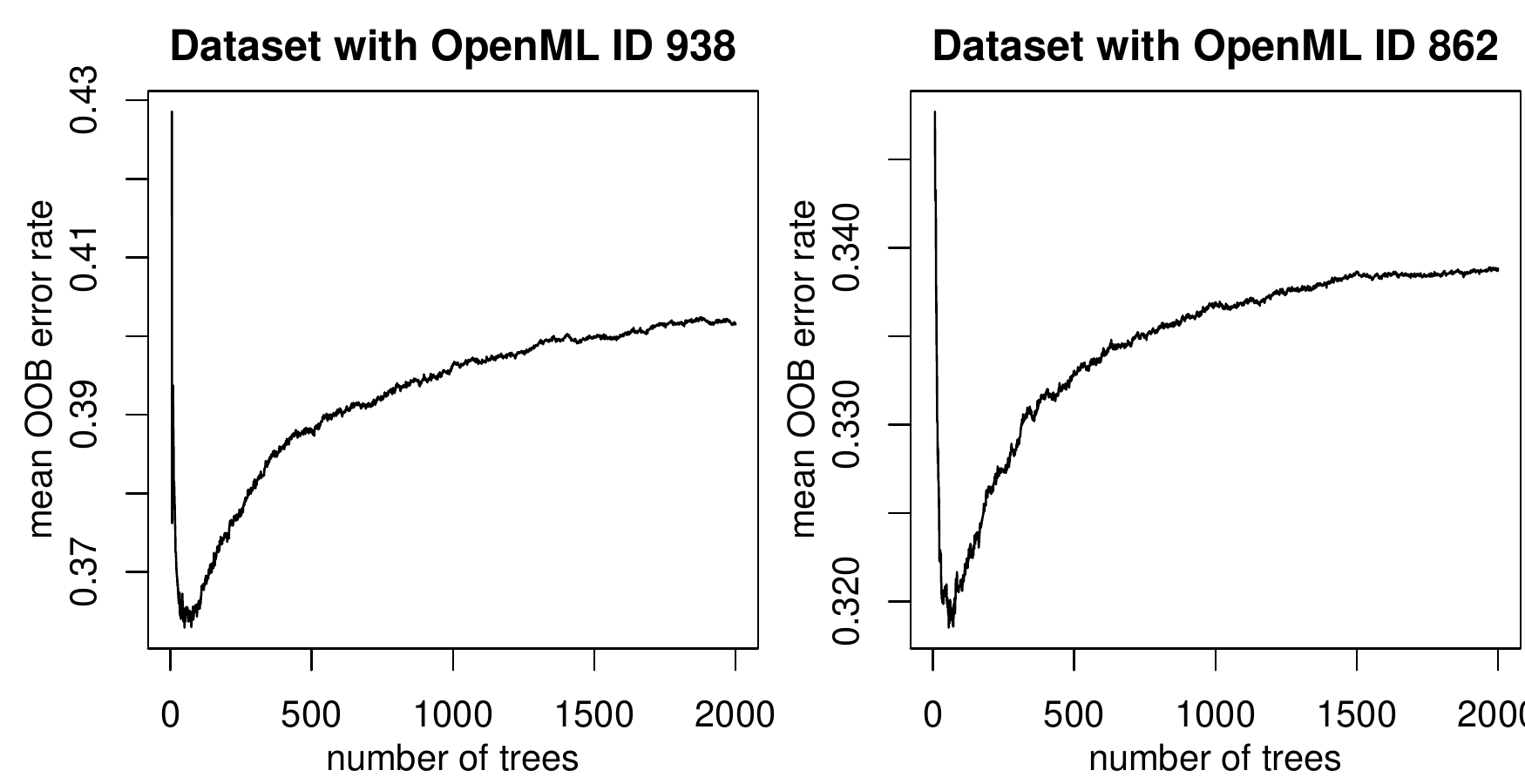}
  \caption{Mean OOB error rate curves for OpenML dataset 938 (left) and OpenML dataset 862 (right). The curves are averaged over 1000 independent runs of random forest with different seeds.}
   \label{fig:2examples}
\end{center}
\end{figure}

To set the scene, we first address this issue empirically by looking at the curve depicting the out-of-bag (OOB) error rate (see Section \ref{sec:background} for a definition of the OOB error) for different number 
of trees (also called OOB error rate curve) for various datasets from the OpenML database \citep{OpenML2013}. 
To obtain more stable results and better estimations for the expected error rate we repeat this procedure 1000 times for each dataset and average the results. 
Interestingly, on most datasets we observe monotonously decreasing curves with growing number of trees, while others yield strange non-monotonous patterns. 
Two of these strange error rate curve examples are depicted in Figure \ref{fig:2examples} as an illustration. It can be seen from these curves that the initial error rate drops steeply before starting to increase after a certain number 
of trees to finally reach a plateau.

At first view, such non-monotonous patterns are a clear argument in favor of tuning  $T$. We claim, however, that it is important to understand why and in which circumstances such patterns happen in order to decide 
whether or not $T$ should be tuned in general. In Section \ref{sec:theory}, we address this issue from a theoretical point of view, by formulating the expected error rate as a function of the probabilities $\varepsilon_i$ 
of correct classification by a single tree for each observation $i$ of the training dataset, for $i=1,\dots,n$ (with $n$ denoting the size of the training dataset). This theoretical view provides a clear explanation of the 
non-monotonous error rate curve patterns in the case of classification. With a similar approach, we show that such non-monotonous patterns cannot be obtained with the Brier score or the logarithmic loss as performance measures, which are based on probability estimations and also not for the mean squared error in the case of regression. Only for the AUC we can see non-monotonous curves as well. 

The rest of this paper is structured as follows. Section \ref{sec:background} is a brief introduction into random forest and performance estimation. Theoretical results are presented in Section \ref{sec:theory}, while the 
results of a large empirical study based on 306 datasets from the public database OpenML \citep{OpenML2013} are reported in Section \ref{sec:empirical}. More precisely, we empirically validate our theoretical model for the 
error as a function of the trees as well as our statements regarding the properties of datasets yielding non-monotonous patterns. We finally argue in Section \ref{sec:discussion} that there is no inconvenience---except 
additional computational cost---in adding trees to a random forest and that $T$ should thus not be seen as a tuning parameter.

\section{Background: random forest and measures of performance}
\label{sec:background}

In this section we will tell some general things about random forest and introduce the general notation and some well known performance measures. 

\subsection{Random forest}
The random forest (RF) is an ensemble learning technique consisting of the aggregation of a large number $T$ of decision trees, resulting in a reduction of variance compared to the single decision trees. 
In this paper we consider the original version of RF first described by \citep{Breiman2001}, while acknowledging that other variants exist, for example RF based on conditional inference trees \citep{Hothorn2006} 
which address the problem of variable selection bias investigated by \citet{Strobl2007}, extremely randomized trees \citep{Geurts2006} or ensembles of optimal trees \citep{Khan2016}. Our considerations are 
however generalizable to many of the available RF variants and other methods that use bagging. 

Each tree of the random forest is built based on a bootstrap sample (or a subsample) drawn randomly from the original training dataset using the CART method and the Gini impurity 
\citep{Breiman1984} as the splitting criterion. In regression the split criterion is the residual sum of squares. During the building of each tree of the forest, at each split, only a given number of candidate variables are considered. 
Random forest is considered a black-box algorithm, as gaining insight on a RF prediction rule is hard due to the large number of trees. One of the most common approaches to extract from 
the random forest interpretable information on the contribution of the different variables consists in the computation of the so-called variable importance. 

A prediction is obtained for a new observation by aggregating the predictions made by the $T$ single trees. In the case of regression RF, the most straightforward and common procedure 
consists of averaging the prediction of the single trees, while majority voting is usually applied to aggregate classification trees. This means that the new observation is assigned to the 
class that was most often predicted by the $T$ trees. 


While RF can be used for various types of response variables including censored survival times or multicategorical variables, in this paper we mainly focus on the two most common cases,  binary classification and regression except if otherwise stated.

\subsection{General notations}
From now on, we consider a fixed training dataset $D$ consisting of $n$ observations, which is used to derive prediction rules by applying the RF algorithm with a number $T$ of trees.
Ideally, the performance of these prediction rules is estimated based on an independent test dataset, denoted as $D_{test}$, consisting of $n_{test}$ test observations.

Considering the $i$th observation from the test dataset ($i=1,\dots,n_{test}$), we denote its true response as $y_i$, which can be either a numeric value (in the case of regression) or the binary label 0 vs. 1 (in the case of binary classification).
The predicted value output by tree $t$ (with $t=1,\dots,T$) is denoted as $\hat{y}_{it}$, while $\hat{y}_i$ stands for the predicted value output by the whole random forest. Note that, in the case of regression, $\hat{y}_i$ is usually obtained by averaging as 
\[
\hat{y}_i=\frac{1}{T}\sum_{t=1}^T\hat{y}_{it}.
\]
In the case of classification, $\hat{y}_i$ is usually obtained by majority voting with a threshold of 0.5. For binary classification, it is equivalent to computing the same average as for regression, which now takes the form
\[
\hat{p}_i=\frac{1}{T}\sum_{t=1}^TI ( \hat{y}_{it}=1 )
\] 
and is denoted as $\hat{p}_i$ (standing for probability), and finally deriving $\hat{y}_i$ as
\[
   \hat{y}_{i} =
   \begin{cases}     1 & \text{if }\hat{p}_i>0.5,  \\
     0 & \text{otherwise.}
   \end{cases}
\]

\subsection{Measures of perfomance for binary classification and regression}
\label{subsec:errormeasures}
In regression as well as in classification, the performance of a RF for observation $i$ is usually quantified through a so-called loss function measuring the discrepancy between the true response $y_i$ and the predicted response $\hat{y}_i$ or, in the case of binary classification, between $y_i$ and $\hat{p}_i$. 

For both regression and binary classification, the classical and most straightforward measure is defined for observation $i$ as 
\[
e_{i}\ =\ (y_i-\hat{y}_{i})^2\ =\ L(y_i,\hat{y}_{i})
\]
with $L(.,.)$ standing for the loss function $L(x,y)=(x-y)^2$.  In the case of regression this is simply the squared error. Another common loss function in the regression case is the absolute loss $L(x,y)=|x-y|$.  For binary classification both measures simplify to
\[
   e_{i} =
   \begin{cases}     0 & \text{if observation } i \text{ is classified correctly by the RF}  \\
     1 & \text{otherwise,}
   \end{cases}
\]
which we will simply denote as {\it error} from now on.
One can also consider the performance of single trees, i.e., the discrepancy between $y_i$ and $\hat{y}_{it}$. We define $e_{it}$ as 
\[
e_{it}=L(y_i,\hat{y}_{it})=(y_i-\hat{y}_{it})^2
\]
and the mean error---a quantity we need to derive our theoretical results on the dependence of performance measures on the number of tree $T$---as
\[
\varepsilon_i = E(e_{it}),
\]
where the expectation is taken over the possible trees conditionally on $D$. 
The term $\varepsilon_i$ can be interpreted as the difficulty to predict $y_i$ with single trees. 
In the case of binary classification, we have $(y_i-\hat{y}_{it})^2=|y_i-\hat{y}_{it}|$ and $\varepsilon_i$ can be simply estimated as $|y_i-\hat{p}_i|$  from a RF with a large number of trees. 

In the case of binary classification, it is also common to quantify performance through the use of the Brier score, which has the form
\[
b_i\ =\ (y_i-\hat{p}_i)^2=L(y_i,\hat{p}_i)
\]
or of the logarithmic loss
\[
l_i\ =\ -(y_i\ln(\hat{p}_i)+(1-y_i)\ln (1-\hat{p}_i)).
\]
Both of them are based on $\hat{p}_i$ rather than $\hat{y}_i$, and can thus be only defined for the whole RF and not for single trees. 

The area under the ROC curve (AUC) cannot be expressed in terms of single observations, as it takes into account all observations at once by ranking the $\hat{p}_i$-values. It can be interpreted 
as the probability that the classifier ranks a randomly chosen observation with $y_i=1$ higher than a randomly chosen observation with $y_i=0$. The larger the AUC, the better the discrimination between the two classes. The (empirical) AUC is defined as
\[
 \text{AUC} = \frac{\sum_{i=1}^{n_1}\sum_{j=1}^{n_2} S(\hat{p}_i^{\star}, \hat{p}_j^{\star \star})}{n_1 n_2},
\]
where $\hat{p}_1^{\star},..., \hat{p}_{n_1}^{\star}$ are probability estimations for the $n_1$ observations with $y_i=1$, $\hat{p}_1^{\star \star},..., \hat{p}_{n_2}^{\star \star}$ probability estimations for the $n_2$ observations with $y_i=0$ and $S(.,.)$ is defined as
\[
S(p, q) =   \begin{cases}  0 & \text{if } p<q  \\
                             0.5 & \text{if } p = q \\
                             1 & \text{if } p>q.
   \end{cases}
\]
The AUC can also be interpreted as the Mann-Whitney U-Statistic divided by the product of $n_1$ and $n_2$.

\subsection{Measures for multiclass classification}
\label{subsec:multiclassmeasures}

The measures defined in the previous section can be extended to the multiclass classification case. 
Let $K$ denote the number of classes ($K>2$). The response  $y_i$ takes values in $\{1,...,K\}$. 
The error for observation $i$ is then defined as 
\[
e_i = I(y_i \neq \hat{y}_i).
\]
We denote the estimated probability of class $k$ for observation $i$ as 
\[
 \hat{p}_{ik} = \frac{1}{T} \sum_{t=1}^T I(\hat{y}_{it}= k).
\]
The logarithmic loss is then defined as 
\[
l_i = \sum_{k=1}^K -I(y_i=k) \log(\hat{p}_{ik}) 
\]
and the generalized Brier score is defined as 
\[
 b_i = \sum_{k=1}^K (\hat{p}_{ik} - I(y_i=k))^2,
\]
which in the binary case is twice the value of the definition that was used in the previous section. 
Following \citet{Hand2001}, the AUC can also be generalized to the multiclass case as  
\[
\text{AUC} = \frac{1}{K(K-1)} \sum_{j=1}^K \sum_{\substack{k=1 \\ k\neq j}}^K \text{AUC}(j,k), 
\]
 where $\text{AUC}(j,k)$ is the AUC between class $k$ and $j$, see also \citet{Ferri2009} for more details. 
It is equivalent to the definition given in Section \ref{subsec:errormeasures} in the binary classification case. 

\subsection{Test dataset error vs. out-of-bag error}
\label{subsec:oobmeasures}
In the cases where a test dataset $D_{test}$ is available, performance can be assessed by averaging the chosen performance measure (as described in the previous paragraphs) over the $n_{test}$ observations.
For example the classical error rate (for binary classification) and the mean squared error (for regression) are computed as
\[
\frac{1}{n_{test}}\sum_{i=1}^{n_{test}} L(y_i,\hat{y}_{i}),
\]
with $L(x,y)=(x-y)^2$, while the mean absolute error (for regression) is obtained by defining $L(.,.)$ as $L(x,y)=|x-y|$.
Instead of averaging it is also possible (in the regression case) to consider the median 
\[
 \mathrm{median}\left(L(y_i,\hat{y}_{1}), ..., L(y_i,\hat{y}_{n_{test}})\right),
\]
of these values, which results in the median squared error for the loss function $L(x,y)=(x-y)^2$ and in the  median absolute error for the loss function $L(x,y)=|x-y|$.

An alternative for using a test dataset is the out-of-bag error which is calculated by using the out-of-bag (OOB) estimations of the training observations. OOB estimations are calculated by predicting the class, 
the probability (in the classification case) or the real value (in the regression case) for each training observation $i$ (for $i=1,\dots,n$) by using only the trees for which this observation was not included in the bootstrap sample, i.e., was not used to construct the tree. 
These estimations are compared to the true values by calculating some performance measure such as those mentioned in Section \ref{subsec:errormeasures}, \ref{subsec:multiclassmeasures} and \ref{subsec:oobmeasures}. 


\newpage
\section{Theoretical results}
\label{sec:theory}
In this section we compute the expected performance---according to the error, the Brier score and the logarithmic loss outlined in Section \ref{subsec:errormeasures}---of 
a binary classification or regression RF consisting of $T$ trees as estimated based on the $n_{test}$ test observations, while considering the training dataset as fixed. 
For the AUC we prove that it can be a non-monotonous function in $T$. The case of other measures and multiclass classification will be investigated empirically in Section \ref{sec:empirical}.

In this section we are concerned with \textit{expected} performances, where expectation is taken over the sets of $T$ trees. Our goal is to study the monotonicity of the expected errors with respect to $T$.
The number $T$ of trees is considered a parameter of the RF and now mentioned in parentheses everytime we refer to the whole forest.

\subsection{Error rate (binary classification)}
We first show that for single observations the OOB error rate curve can be increasing and then show exemplified how this 
can influence the shape of the average curve of several observations. 
\subsubsection{Theoretical considerations}
Let us first consider the classical error rate $e_i(T)$ for observation $i$ with a RF including $T$ trees and derive its expectation, conditionally on the training set $D$. 
\begin{eqnarray*}
 E(e_i(T)) & = &  E \left(I \left( \frac{1}{T} \sum_{t=1}^T e_{it}  > 0.5 \right) \right)\\
 & = & P \left( \sum_{t=1}^T e_{it}  > 0.5 \cdot T \right). 
\end{eqnarray*}
We note that $e_{it}$ is a binary variable with $E(e_{it})=\varepsilon_i$. Given a fixed training dataset $D$ and observation $i$, the $e_{it}$, $t=1,...,T$ are mutually independent. It follows that  the sum $X_i=\sum_{t}^Te_{it}$ 
follows the binomial distribution $B(T, \varepsilon_i)$. 
It is immediate that the contribution of observation $i$ to the expected error rate, $P(X_i>0.5 \cdot T)$, is an increasing function in $T$ for $\varepsilon_i > 0.5$ and a decreasing function in $T$ for $\varepsilon_i < 0.5$. 

Note that so far we ignored the case where $\sum_{t=1}^T e_{it}  = 0.5 \cdot T$, which may happen when $T$ is even. 
In this case, the standard implementation in R (\texttt{randomForest}) assigns the observation randomly 
to one of the two classes. This implies that the term $0.5 \cdot P(\sum_{t=1}^T e_{it}  = 0.5 \cdot T)$ has to be added to the above term, which does not affect our considerations on the role of $\varepsilon_i$.

\subsubsection{Impact on error rate curves}
\begin{figure}[!htb]
\begin{center}
  \includegraphics[width=\textwidth]{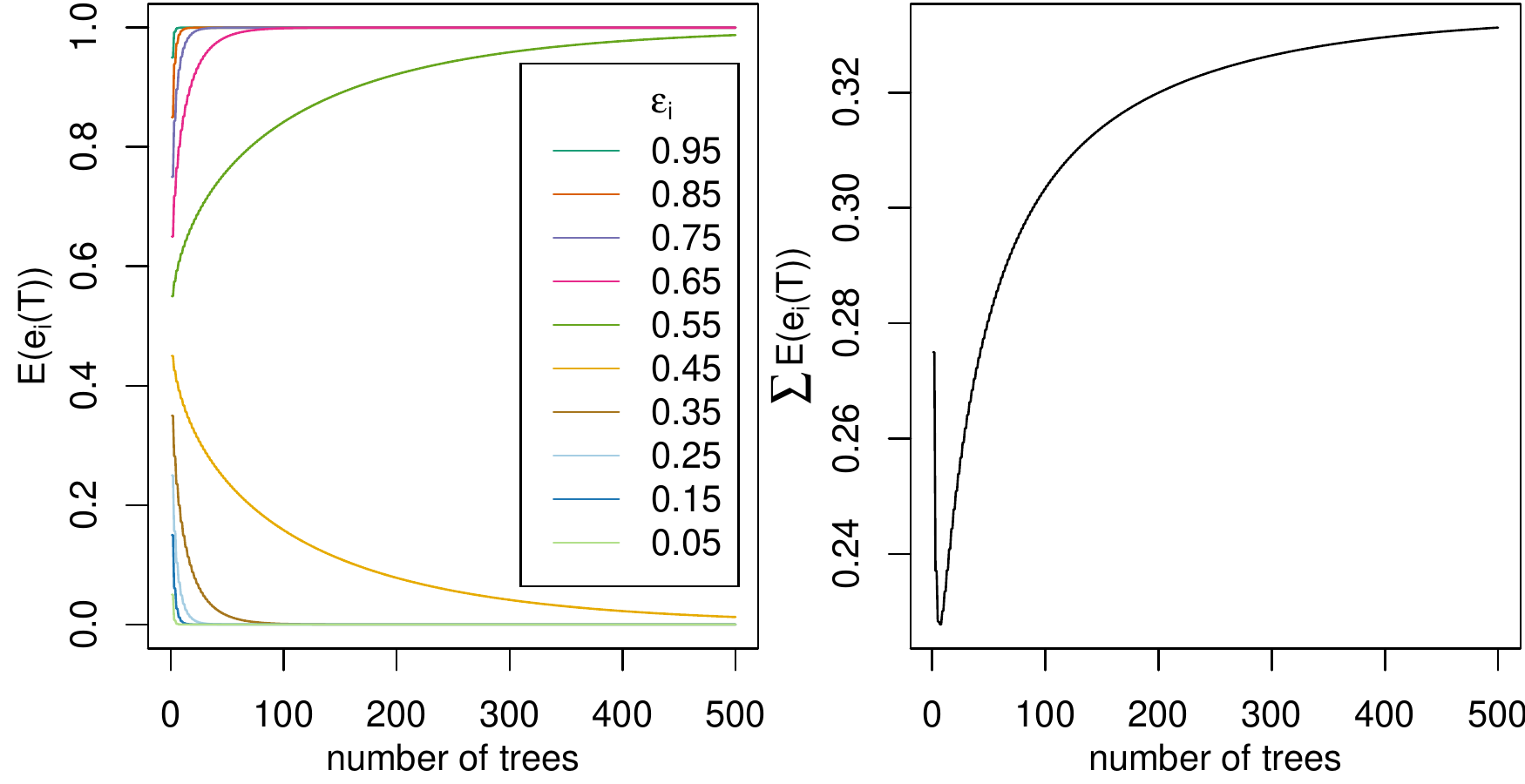}
  \caption{Left: Expected error rate curves for different $\varepsilon_i$ values // Right: Plot of the average curve of the curves with $\varepsilon_1 = 0.05$, $\varepsilon_2 = 0.1$, $\varepsilon_3 = 0.15$, 
  $\varepsilon_4 = 0.2$, $\varepsilon_5 = 0.55$ and $\varepsilon_6 = 0.6$}
  \label{fig:errorcurves}
\end{center}
\end{figure}

The error rate curve for observation $i$ is defined as the curve described by the function $e_i: T \rightarrow \mathbb{R}$.
The expectation $ E(e_i(T))$ of the error rate curve for observation $i$ with the mentioned adjustment in the case of an even number of trees  can be seen in the left plot of Figure \ref{fig:errorcurves} for different values of $\varepsilon_i$. 
Very high and very low values of $\varepsilon_i$ lead to rapid convergence, while for $\varepsilon_i$-values close to 0.5 more trees are needed to reach the plateau. 

The error rate curve obtained for a test dataset consists of the average of the error rate curves of the single observations. Of course, if trees are good classifiers we should have $\varepsilon_i < 0.5$ for most observations. 
In many cases, observations with  $\varepsilon_i > 0.5$ will be compensated by observations with $\varepsilon_i < 0.5$ in such a way that the expected error rate curve is monotonously decreasing. 
This is typically the case if there are many observations with $\varepsilon_i\approx 0$ and a few with $\varepsilon_i\approx 1$.

However, if there are many observations with  $\varepsilon_i\approx 0$ and a few  observations with $\varepsilon_i \geq 0.5$ that are close to 0.5, the expected error rate curve initially
falls down quickly because of the observation with $\varepsilon_i \approx 0$ and then grows again slowly as the number of trees increases because of the observations with $\varepsilon_i \geq 0.5$ close to 0.5. 
In the right plot of Figure \ref{fig:errorcurves} we can see the mean of the expected error rate curves for $\varepsilon_1 = 0.05$, $\varepsilon_2 = 0.1$, $\varepsilon_3 = 0.15$, 
  $\varepsilon_4 = 0.2$, $\varepsilon_5 = 0.55$ and $\varepsilon_6 = 0.6$ and can see exactly the pattern that we expected.
In Section \ref{sec:empirical} we will see that the two example datasets whose non-monotonous out-of-bag error rate curves are depicted in the introduction have a similar distribution of $\varepsilon_i$. 

We see that the convergence rate of the error rate curve is only dependent on the distribution of the $\varepsilon_i$ of the observations. 
Hence, the convergence rate of the error rate curve is not directly dependent on the number of observations $n$ or the number of features, 
but these characteristics could influence the empirical distribution of the $\varepsilon_i$'s and hence possibly the convergence rate as outlined in Section \ref{subsubsec:overall}.  

\subsection{Brier Score (binary classification) and squared error (regression)}
We now turn to the Brier score and compute the expected Brier score contribution of observation $i$ for a RF including $T$ trees, conditional on the training set $D$. We obtain
\begin{eqnarray*}
 E(b_i(T)) & = & E((y_i - \hat{p}_i(T) ) ^2) \\
                    & = & E \left( \left( y_i-\frac{1}{T}\sum_{t=1}^T\hat{y}_{it} \right)^2 \right) \\
                    & = & E \left( \left( \frac{1}{T}\sum_{t=1}^T(y_i-\hat{y}_{it}) \right)^2 \right) \\
                    & = & E \left( \left( \frac{1}{T}\sum_{t=1}^T e_{it} \right)^2 \right).
\end{eqnarray*}

From $E(Z^2) = E(Z)^2 + Var(Z)$ with $Z=\frac{1}{T}\sum_{t=1}^T e_{it}$  it follows:
\begin{eqnarray*}
 E(b_i(T)) & = & E(e_{it})^2+\frac{Var(e_{it})}{T},
\end{eqnarray*} 
which is obviously a strictly monotonous decreasing function of $T$. This also holds for the average over the observations of the test dataset. 
In the case of binary classification, we have  $e_{it}\sim\mathcal{B}(1,\varepsilon_i)$, yielding $E(e_{it})=\varepsilon_i$ and $Var(e_{it})=\varepsilon_i(1-\varepsilon_i)$, thus allowing the formulation of  
$E(b_i(T))$ as $E(b_i(T))=\varepsilon_i^2+\frac{\varepsilon_i(1-\varepsilon_i)}{T}$.

Note that the formula $ E(b_i(T)) =  E(e_{it})^2+Var(e_{it})/T$ is also valid for the squared error in the regression case, except that in this case we would write $\hat{y}_{i}$ instead of $\hat{p}_{i}$ in the first line. 

\subsection{Logarithmic Loss (binary classification)}
As outlined in Section \ref{subsec:errormeasures}, another usual performance measure based on the discrepancy between $y_i$ and $\hat{p}_i$ is the logarithmic loss $l_i(T)=-(y_i\ln (\hat{p}_i(T))+(1-y_i)\ln (1-\hat{p}_i(T)))$. 
Noticing that $\hat{p}_i(T)=1-\frac{1}{T}\sum_{t=1}^Te_{it}$ for $y_i=1$ and $\hat{p}_i(T)=\frac{1}{T}\sum_{t=1}^Te_{it}$ for $y_i=0$, it can be in both cases $y_i=0$ and $y_i=1$ reformulated as
\[
l_i(T) =  -\ln \left( 1-\frac{1}{T}\sum_{t=1}^T e_{it} \right).
\]
In the following we ensure that the term inside the logarithm is never zero by adding a very small value $a$ to $1-\frac{1}{T}\sum_{t=1}^T e_{it}$. The logarithmic loss $l_i(T)$ is then always defined and its expectation exists. 
This is similar to the solution adopted in the \texttt{mlr} package, where $10^{-15}$ is added in case that the inner term of the logarithm equals zero. 

With $Z:=  1 - \frac{1}{T} \sum_{t=1}^T e_{it} + a$,  we can use the Taylor expansion \citep{Casella2002},
\begin{align}
E\left[f(Z)\right]  {} = & E\left[f(\mu_Z + \left(Z - \mu_Z\right))\right] \nonumber \\
\approx & E \left[f(\mu_Z) + f'(\mu_Z)\left(Z-\mu_Z\right) + \frac{1}{2}f''(\mu_Z) \left(Z - \mu_Z\right)^2 \right] \nonumber \\
 = & f(\mu_Z) + \frac{f''(\mu_Z)}{2} \cdot Var(Z) \nonumber \\
 = & f(E(Z)) + \frac{f''(E(Z))}{2} \cdot Var(Z) \nonumber,
\end{align}
where $\mu_Z$ stands for $E(Z)$ and $f(.)$ as $f(.)=-\ln (.)$.
We have
\begin{align}
Var(Z) & =  \frac{\varepsilon_i(1-\varepsilon_i)}{T} \nonumber\\
E(Z) & = 1 - \varepsilon_i + a \nonumber\\
f(E(Z)) & =  -\ln(1 - \varepsilon_i + a) \nonumber\\
f''(E(Z)) & =  (1 - \varepsilon_i + a)^{-2} \nonumber,
\end{align}
finally yielding
\[
E(l_i(T)) \approx  -\ln (1 - \varepsilon_i + a)+\frac{\varepsilon_i (1- \varepsilon_i)}{2T (1 - \varepsilon_i + a)^2},
\]
which is obviously a decreasing function of $T$. The Taylor approximation gets better and better for increasing $T$, since the variance of $l_i(T)$ decreases with increasing $T$ and thus $l_i(T)$ 
tends to get closer to its expectancy. 

\subsection{Area Under Curve (AUC)}
For the AUC, considerations such as those we made for the error rate, the Brier score and the logarithmic loss are impossible, since the AUC is not the sum of individual contributions of the observations. It is however relatively easy to see that the expected AUC is not always an increasing function of the number $T$ of trees. For example, think of the trivial example of a test dataset consisting of two observations with responses $y_1$ resp. $y_2$ and $E(\hat{p}_1(T))=0.4$ resp. $E(\hat{p}_2(T))=0.6$.  
If $y_1 = 0$ and $y_2 = 1$, the expected AUC curve increases monotonously with $T$, as the probability of a correct ordering according to the calculated scores $\hat{p}_1(T)$ and $\hat{p}_2(T)$ increases. However, if $y_1 = 1$ and $y_2 = 0$, we obtain a monotonously decreasing function, as the probability of a wrong ordering gets higher with increasing number of trees. 
It is easy to imagine that, for different combinations of $E(\hat{p}_i(T))$, one can obtain increasing curves, decreasing curves or non-monotonous curves.

\subsection{Adapting the models to the OOB error}

If we consider the OOB error instead of the test error from an independent dataset, the formulas given in the previous subsections are not directly applicable. After having trained $T$ trees, for making an OOB estimation for an observation we can only use the trees for which the observation was out-of-bag. If we take a simple bootstrap sample from the $n$ training observation when bagging we have {\it on average} only $T \cdot (1-\frac{1}{n})^n \approx T\cdot \exp{(-1)} \approx T \cdot 0.367$ trees for predicting the considered observation. 
This means that we would have to replace $T$ by $T \cdot \exp{(-1)}$ in the above formulas and that the formulas are no longer exact because $T\cdot\exp{(-1)}$  is only an average. Nonetheless it is still a good approximation as confirmed in our benchmark experiments. 

\section{Empirical results}
\label{sec:empirical}

This section shows a large-scale empirical study based on 193 classification tasks and 113 regression tasks from the public database OpenML \citep{OpenML2013}. 
The datasets were downloaded with the help of the \texttt{OpenML} R package \citep{Casalicchio2017}.
The goals of this study are to 
(i) give an order of magnitude of the frequency of non-monotonous patterns of the error rate curve in real data settings; 
(ii) empirically confirm our statement that observations with $\varepsilon_i$ greater than (but close to) 0.5 are responsible for non-monotonous patterns;
(iii) analyse the results for other classification measures, the multiclass classification and several regression measures;
(iv) analyse the convergence rate of the OOB curves.

\subsection{Study design}

We select the 193 classification tasks and 113 regression tasks from the OpenML platform \citep{OpenML2013} satisfying the following criteria: 
the datasets have predefined tasks in OpenML (see \citet{OpenML2013} for details on the OpenML nomenclature) and they include less than 1000 observations  and less than 1000 features, in order to keep the computation time for training RF feasible.
Cleaning procedures such as the deletion of duplicated datasets are also applied to obtain a decent collection of datasets. 
From the 193 classification tasks, 149 are binary classification tasks and 44 multiclass classification tasks. 

For each dataset we run the RF algorithm with $T=2000$ trees 1000 times successively with different seeds using the R package \texttt{randomForest} \citep{Liaw2002} with the default parameter settings. We choose 2000 trees because in a preliminary study on a subset of the datasets we could 
observe that convergence of the OOB curves was reached within these 2000 trees.

For the classification tasks we calculate the OOB curves for the error rate, the balanced error rate, the (multiclass) Brier score, the logarithmic loss and the (multiclass) AUC using our new package \texttt{OOBCurve}, see details in the next section. 

For the regression tasks we calculate the OOB curves using the mean squared error, the mean absolute error, the median squared error and the median of absolute error as performance measures. We parallelize the computations using the R package 
\texttt{batchtools} \citep{Bischl2015}. 
For each measure and each dataset, the final curve is obtained by simply averaging over the 1000 runs of RF. We plot each of them in three files separetely for binary classification, multiclass classification and regression. 
In the plots the x-axis starts at $T=11$ since overall performance estimates are only defined if each observation was out-of-bag in at least one of the $T$ trees, which is not always the case in practice for $T<10$. 
We plot the curves only until $T=500$, as no interesting patterns can be observed after this number of trees (data not shown). 
The graphics, the R-codes and the results of the benchmark experiment can be found on \url{https://github.com/PhilippPro/tuneNtree}. 

\subsection{The R package \texttt{OOBCurve}}
The calculation of out-of-bag estimates for different performance measures is implemented in our new R package \texttt{OOBCurve}. More precisely, it takes a random forest constructed with the R package \texttt{randomForest} \citep{Liaw2002} or \texttt{ranger} \citep{Wright2016} as input 
and can calculate the OOB curve for any measure that is available from the \texttt{mlr} 
package \citep{Bischl2016}. See our R-code for application examples. The \texttt{OOBCurve} package will soon be available on CRAN R package repository and is already available on Github 
(\url{https://github.com/PhilippPro/OOBCurve}).

\subsection{Results for binary classification}

The average gain in performance in the out-of-bag performance for 2000 trees instead of 11 trees is -0.0324 for the error rate, -0.0683 for the brier score, -2.383 for the logarithmic loss and 0.0553 for the AUC. 
In the following we will concentrate on the visual analysis of the graphs and are especially interested in the results of the error rate.  

\subsubsection{Overall results for the OOB error rate curves}
\label{subsubsec:overall}
We observe in the graphs of the OOB error rate curves that for most datasets the curve is quickly decreasing  until it converges to a dataset-specific plateau value. 
In 15 cases which make approximately $10\%$ of the datasets, however, the curve grows again after reaching its lowest value, leading to a value at 2000 trees that is by at least 0.005 bigger 
than the lowest value of the OOB error rate curve for $T \in [10,150]$. 
This happens mainly for smaller datasets, where a few observations can have a high impact on the error curve. Of these 15 cases 14 belong to the smaller half of the datasets---ordered by the number of observations multiplied with the number of features. 

\subsubsection{Datasets with non-monotonous OOB error rate curve}
We now examine in more detail the datasets yielding non-monotonous patterns. In particular, the histograms of the estimates $\hat{\varepsilon}_i=|y_i-\hat{p}_i|$ of the observation-specific errors 
$\varepsilon_i$  are of interest, since our theoretical results prove that the distribution of the $\varepsilon_i$ determines the form of the expected error rate curve. 
To get these histograms we compute the estimates $\hat{\varepsilon}_i$  of the observation-specific errors 
$\varepsilon_i$ (as defined in Section \ref{subsec:errormeasures}) from a RF with a huge number $T=100000$: the more trees, the more accurate the estimates of $\varepsilon_i$. 

\begin{figure}[!htb]
\begin{center}
  \includegraphics[width=\textwidth]{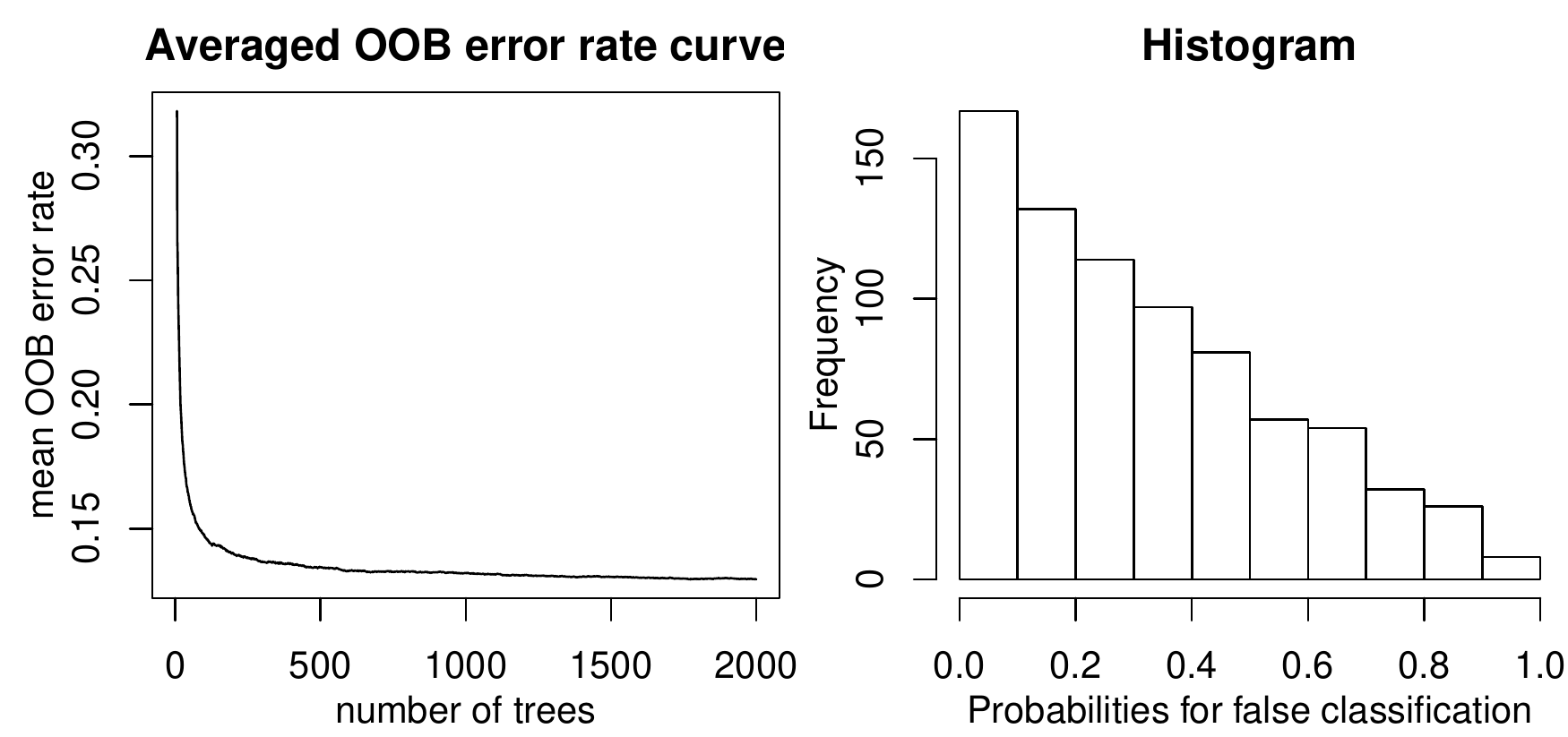}
  \caption{Dataset with OpenML ID 37: Mean OOB error rate curve over 1000 runs of random forests (left) and histogram of the estimates of $\varepsilon_i$ ($i=1,\dots,n$) from a random forest with 100000 trees}
  \label{fig:3OOBhist}
\end{center}
\begin{center}
  \includegraphics[width=\textwidth]{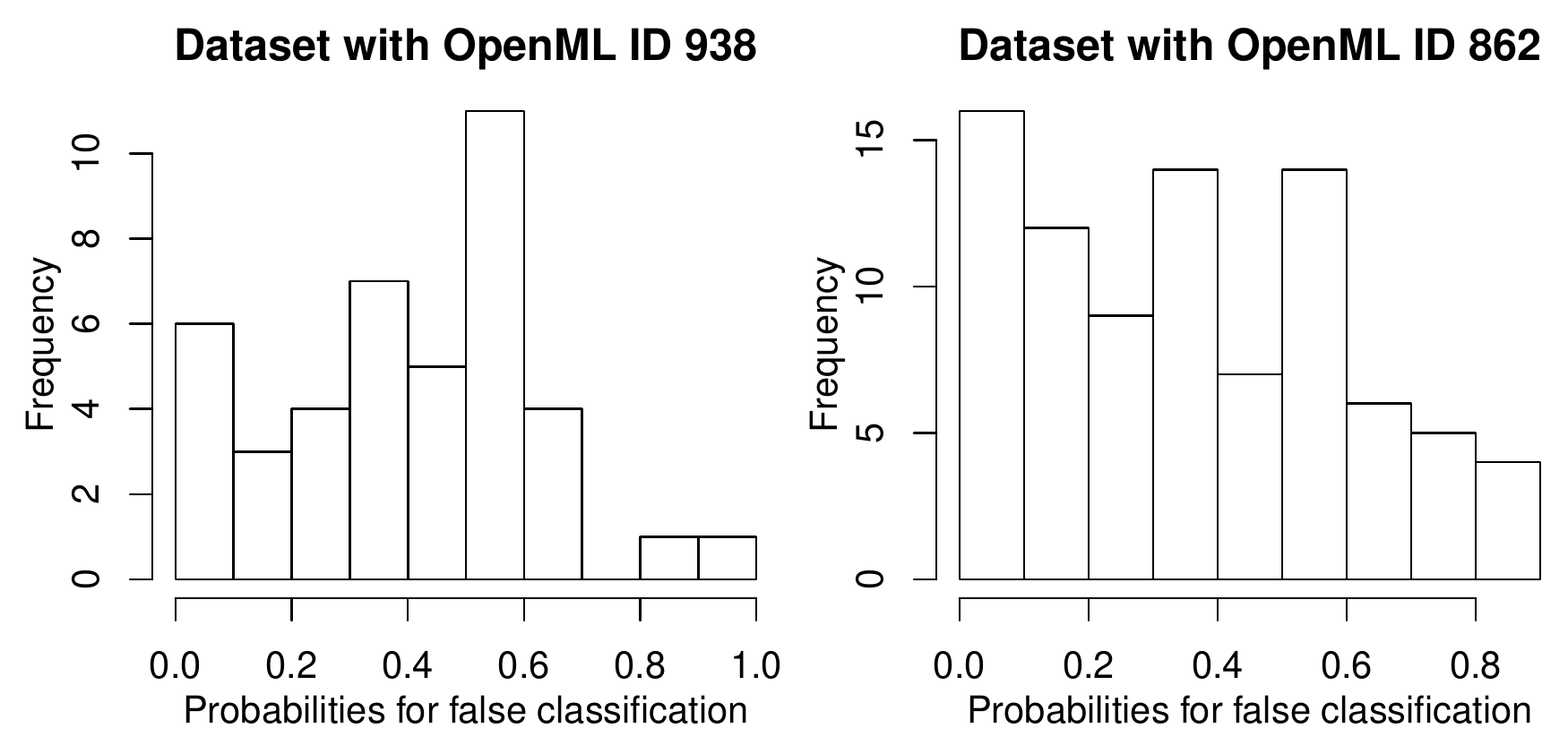}
  \caption{Histogram of the estimates of $\varepsilon_i$ ($i=1,\dots,n$) from random forests with 100000 trees for dataset 938 (left) and dataset 862 (right)}
  \label{fig:4hist}
\end{center}
\end{figure}

Such a histogram is displayed in Figure \ref{fig:3OOBhist}. The average OOB error rate curve (left panel) shows the typical pattern of a quickly decreasing OOB error rate. The corresponding histogram of the $\varepsilon_i$ is shown in the right panel. Most $\hat{\varepsilon}_i$ are smaller than 0.5, yielding a monotonously decreasing error rate curve.

Representative examples of histograms in the case of non-monotonous error rate curves can be seen in Figure  \ref{fig:4hist}. The histograms belong to the datasets with OpenML ID 938 and 862 for which the OOB error rate curve was already depicted in the introduction in Figure \ref{fig:2examples}. 
In both cases we see that a non-negligible proportion of observations have $\varepsilon_i$ larger than but close to 0.5. This is in agreement with our theoretical results. With growing number of trees the chance that these observations are incorrectly classified increases, while the chance for observations with 
$\varepsilon_i\approx 0$ is already very low--- and thus almost constant. 

\subsubsection{Other measures}

For the Brier score and the logarithmic loss we observe, as expected, monotonically decreasing curves for all datasets. 
The expected AUC curve usually appears as growing function in $T$. In a few datasets such as the third binary classification example (OpenML ID 905), however, it falls after reaching a maximum. 

To measure the relationship between the different curves, we calculated the Bravais-Pearson linear correlation and Kendall's $\tau$ rank correlation between the values of the OOB curves of the different performance 
measures and averaged these correlation matrices over all datasets. 
Note that we do not perform any correlation tests, since the assumption of independent identically distributed observations required by these tests is not fulfilled: our correlation analyses are meant to be explorative. 
The results can be seen in Table \ref{tab:cor_bin}. The Brier score and logarithmic loss have the highest correlation. They are also more correlated to the AUC than to 
the error rate, which has the lowest correlation to all other measures. 

\begin{table}[ht]
\centering
\begin{tabular}{rrrrr}
  \hline
 & error rate & Brier score & logarithmic loss & AUC \\ 
  \hline
error rate & 1.00 & 0.28 & 0.27 & -0.18 \\ 
  Brier score & 0.72 & 1.00 & 0.96 & -0.63 \\ 
  logarithmic loss & 0.65 & 0.93 & 1.00 & -0.63 \\ 
  AUC & -0.64 & -0.84 & -0.81 & 1.00 \\ 
   \hline
\end{tabular}
\caption{Linear (bottom-left) and rank (top-right) correlation results for binary classification datasets} 
\label{tab:cor_bin}
\quad
\centering
\begin{tabular}{rrrrr}
  \hline
 & error rate & Brier score & logarithmic loss & AUC \\ 
  \hline
error rate & 1.00 & 0.44 & 0.45 & -0.43 \\ 
  Brier score & 0.86 & 1.00 & 0.98 & -0.87 \\ 
  logarithmic loss & 0.84 & 0.95 & 1.00 & -0.87 \\ 
  AUC & -0.85 & -0.95 & -0.92 & 1.00 \\ 
   \hline
\end{tabular}
\caption{Linear (bottom-left) and rank (top-right) correlation results for multiclass classification datasets} 
\label{tab:cor_multi}
\end{table}

\subsection{Results for multiclass classification}
The average gain in performance in the out-of-bag performance for 2000 trees instead of 11 trees is -0.0753 for the error rate, -0.1282 for the brier score, -5.3486 for the logarithmic loss and  0.0723 for the AUC. 
These values are higher than the ones from binary classification. 
The visual observations we made for the binary classification also hold for the multiclass classification. The results for the correlation are quite similar, although the correlation 
(see Table \ref{tab:cor_multi}) is in general higher than in the binary case. 

\subsection{Results for regression}
The average performance gain regarding the out-of-bag performance of the R$^2$ for 2000 trees compared to 11 trees is 0.1249.
In the OOB curves for regression we can observe the monotonously decreasing pattern expected from theory in the case of the most widely used mean squared error (mse). The mean absolute error (mae) seems also to be strictly
decreasing. In contrast, for the median squared error (medse) and the median absolute error (medae), we observe  cases where the curve grows with $T$ either in the small $T$ or in the large $T$ region, although in most of the cases using more trees leads to better results.
The correlation coefficients between the measures can be seen in Table \ref{tab:cor_reg}.

\begin{table}[ht]
\centering
\begin{tabular}{rrrrr}
  \hline
 & mse & mae & medse & medae \\ 
  \hline
mse & 1.00 & 0.88 & 0.20 & 0.17 \\ 
  mae & 0.96 & 1.00 & 0.19 & 0.17 \\ 
  medse & 0.49 & 0.50 & 1.00 & 0.97 \\ 
  medae & 0.43 & 0.45 & 0.99 & 1.00 \\ 
   \hline
\end{tabular}
\caption{Linear (bottom-left) and rank (top-right) correlation results for all dataset} 
\label{tab:cor_reg}
\end{table}

We observe a high correlation between the measures which take the mean of the losses of all observations and between the measures which take the median of the losses. 
Correlation between these two groups of measures are around 0.5 for the linear correlation coefficient and around 0.2 for the rank correlation coefficient. 

\subsection{Convergence}
\label{subsec:convergence}
It is clearly visible in the graphs (\url{https://github.com/PhilippPro/tuneNtree/tree/master/graphics}) that increasing the number of trees from 10 to 500 yields a substantial performance gain in most of the cases. 
In most datasets the biggest gain of performance is reached within the first 250 trees. 
In all datasets and for all measures, at 250 trees at most the curve reaches a value that cannot be improved a lot by adding more trees. 

It is important to remember that for the calculation of the OOB error curve at $T$ only $\exp(-1) \cdot T$ trees where used. Thus, as far as future independent data is concerned, the convergence of the performances is by $\exp(1)\approx 2.718$ 
faster than observed from our OOB curves. 
Having this in mind, our observations are in agreement with the results of \citet{Oshiro2012}, which conclude that after growing 128 trees no big gain in the AUC performance could be achieved by growing more trees. 
Admittedly, we only examined small datasets with number of observations and number of features smaller than 1000, so we did not investigate, if our results are generalizable to bigger datasets as mentioned in their paper.

For the assessment of the convergence rate we generally recommend using measures other than the error rate, such as the Brier score or the logarithmic loss.
Their convergence rate is not so dependent on observations with $\varepsilon_i$ close to 0.5 (in the binary classification case), 
and they give an indication of the general stability of the probability estimations of all observations. 
This can be especially important if the threshold for classification is not set prior to 0.5. 
The new \texttt{OOBCurve} R package is a tool to examine the rate of convergence of the trained random forest with any measure that is available in \texttt{mlr}. 

\section{Discussion and extensions}
\label{sec:discussion}
In this section we first discuss the main results of this paper and finally think about possible extensions of our work. 

\subsection{Why more trees are better}
Non-monotonous expected error rate curves observed in the case of binary classification might be seen as an argument in favour of tuning the number $T$ of trees. Based on our results, we argue, however, that tuning is not recommendable. 

Firstly, non-monotonous patterns are observed only with some performance measures like the error rate and the AUC in case of classification. Measures such as the Brier score or the logarithmic loss, which are based on probabilities rather than on the predicted 
class and can thus be seen as more refined, do not yield non-monotonous patterns, as theoretically proved in Section \ref{sec:theory} and empirically observed based on a very large number of datasets in Section \ref{sec:empirical}. 

Secondly, non-monotonous patterns in the expected error rate curves are the result of a particular rare combination of $\varepsilon_i$'s in the training data. Especially if the training dataset is small, 
the chance is high that the distribution of the $\varepsilon_i$ will be different for independent test data, for example values of $\varepsilon_i$ close to but larger than 0.5 may not be present. 
In this case, the expected error rate curve for this independent future dataset would not be non-monotonous, and a large $T$ is better. 

Thirdly, even in the case of non-monotonous expected error rate curves, the minimal error rate value is usually only slightly (by less than 0.01) smaller than the value at convergence. We argue that 
this very small gain---which, as outlined above, is relevant only for future observations with $\varepsilon_i>0.5$---probably does not compensate the advantage of using more trees in terms of other performance measures or in terms of the
precision of the variable importance measures, which are very commonly used in practice.

A term that is commonly used in relation to ensemble methods and tuning parameters is {\it overfitting}. In boosting \citep{Friedman2001}, including too many trees yields overfitting. A natural question is whether the occasional error rate increase for large $T$ discussed in this paper in the case of RF can be seen as a form of overfitting.  
We claim on the contrary that in the non-monotonous patterns overfitting occurs
for smaller $T$ values, in the sense that the minimal value reached for a small $T$ is the result of the very specific combination of $\varepsilon_i$'s in the training set.

\subsection{Extensions}
Note that our theoretical results are in principle generalizable to most methods based on bagging, since the fact that the base learners are trees does not play any role in our proofs. 
Our theoretical results could possibly be generalised to the multiclass case, as supported by our results obtained with 44 multiclass datasets. More research is needed.

Although we claim that increasing the number of trees cannot harm, our empirical results show that, for most of the examined datasets, the biggest performance gain is achieved when training the first 100 trees. 
However, the rate of convergence may be influenced by other hyperparameters of the RF. For example lower sample size while taking bootstrap samples for each tree, 
bigger constraints on the tree depth or more variables lead to less correlated trees and hence more trees are needed to reach convergence. 

One could also think of an automatic break criterion which stops the training automatically according to the convergence of the OOB curves. 
If variable importance measures are computed, it may be recommended to also consider their convergence. This issue also requires more research.


\acks{
We would like to thank Alexander D{\"u}rre for useful comments on the approximation of the logarithmic loss and Jenny Lee for language editing. 
}

\vskip 0.2in
\bibliography{Ntree_RandomForest}

\newpage
\section*{Supplementary File}
\section*{Online discussions}

\begin{itemize}
\item Is the out-of-bag error the optimal parameter to find the right number of trees? 
\url{http://stackoverflow.com/questions/18541923/what-is-out-of-bag-error-in-random-forests}
\item What parameters to tune in random forest? \url{http://stats.stackexchange.com/questions/53240/practical-questions-on-tuning-random-forests}
\item Does the optimal number of trees in a random forest depend on the number of predictors? 
\url{http://stats.stackexchange.com/questions/36165/does-the-optimal-number-of-trees-in-a-random-forest-depend-on-the-number-of-pred/36183#36183}
\item Should random forests based on same data but different random seeds be com-
pared? 
\url{http://stats.stackexchange.com/questions/222279/should-random-forests-based-on-same-data-but-different-random-seeds-be-compared/222771#222771}
\item Is the number of trees a tuning parameter? 
\url{http://stats.stackexchange.com/questions/50210/caret-and-randomforest-number-of-trees}
\item Does random forest overfit? 
\url{http://stats.stackexchange.com/questions/183973/how-not-to-overfit-a-random-forest-in-r}, 
\url{http://stats.stackexchange.com/questions/144305/should-you-tune-ntree-in-the-random-forest-algorithm}
\item Evaluate random forest with OOB or with cross validation? 
\url{http://stats.stackexchange.com/questions/198839/evaluate-random-forest-oob-vs-cv/199201#199201}
\item Does the out-of-bag error increase with number of trees? 
\url{http://stats.stackexchange.com/questions/183569/random-forest-out-of-bag-error-increases-with-number-of-trees}
\item How is the relation between the tree depth and the number of trees? 
\url{http://stackoverflow.com/questions/34997134/random-forest-tuning-tree-depth-and-number-of-trees2}
\item How to determine the number of trees? 
\url{https://www.researchgate.net/post/How_to_determine_the_number_of_trees_to_be_generated_in_Random_Forest_algorithm}
\end{itemize}

\end{document}